\documentclass[10pt,twocolumn,letterpaper]{article}

\usepackage[pagenumbers]{iccv} 
\usepackage{multirow}
\usepackage{algorithm} 
\usepackage{algorithmic}
\usepackage[symbol]{footmisc}
\definecolor{iccvblue}{rgb}{0.21,0.49,0.74}
\usepackage[pagebackref,breaklinks,colorlinks,allcolors=iccvblue]{hyperref}

\title{Tokenize Image as a Set}

\author{Zigang Geng$^{1,2}$, Mengde Xu$^{2}$, Han Hu$^{2}$, Shuyang Gu$^{2}\footnotemark[2]$\\
$^{1}$University of Science and Technology of China\ \ \ $^{2}$Tencent Hunyuan Research\\
{\tt\small zigang@mail.ustc.edu.cn, cientgu@tencent.com}
}

\begin{document}
\maketitle

\renewcommand{\thefootnote}{\fnsymbol{footnote}}  
\footnotetext[2]{Corresponding Author.}

\begin{abstract}
This paper proposes a fundamentally new paradigm for image generation through set-based tokenization and distribution modeling. Unlike conventional methods that serialize images into fixed-position latent codes with a uniform compression ratio, we introduce an unordered token set representation to dynamically allocate coding capacity based on regional semantic complexity. This TokenSet enhances global context aggregation and improves robustness against local perturbations. To address the critical challenge of modeling discrete sets, we devise a dual transformation mechanism that bijectively converts sets into fixed-length integer sequences with summation constraints. Further, we propose Fixed-Sum Discrete Diffusion—the first framework to simultaneously handle discrete values, fixed sequence length, and summation invariance—enabling effective set distribution modeling. Experiments demonstrate our method's superiority in semantic-aware representation and generation quality. Our innovations, spanning novel representation and modeling strategies, advance visual generation beyond traditional sequential token paradigms. Our code and models are publicly available at \url{https://github.com/Gengzigang/TokenSet}.
\end{abstract}    
\section{Introduction}
\label{sec:intro}

Contemporary visual generation frameworks~\cite{vqgan21patrick,sun2024llamagen,mar24li,titok24yu} predominantly adopt a two-stage paradigm: first compressing visual signals into latent representations, then modeling the low-dimensional distributions. Conventional tokenization methods~\cite{kingma2013auto,vqvae17aaron,vqgan21patrick,vitvqgan22yu} typically employ uniform spatial compression ratios, generating serialized codes with fixed positional correspondence. Consider a beach photo where the upper half is a sky region that contains minimal detail; the lower half contains a semantically dense foreground—current approaches allocate the same number of codes to both regions. This raises a fundamental question: Should visually simplistic regions receive the same representational capacity as semantically rich areas?

\begin{figure}[tp]
    \centering
    \includegraphics[width=\linewidth, trim={8 2 8 0},clip]{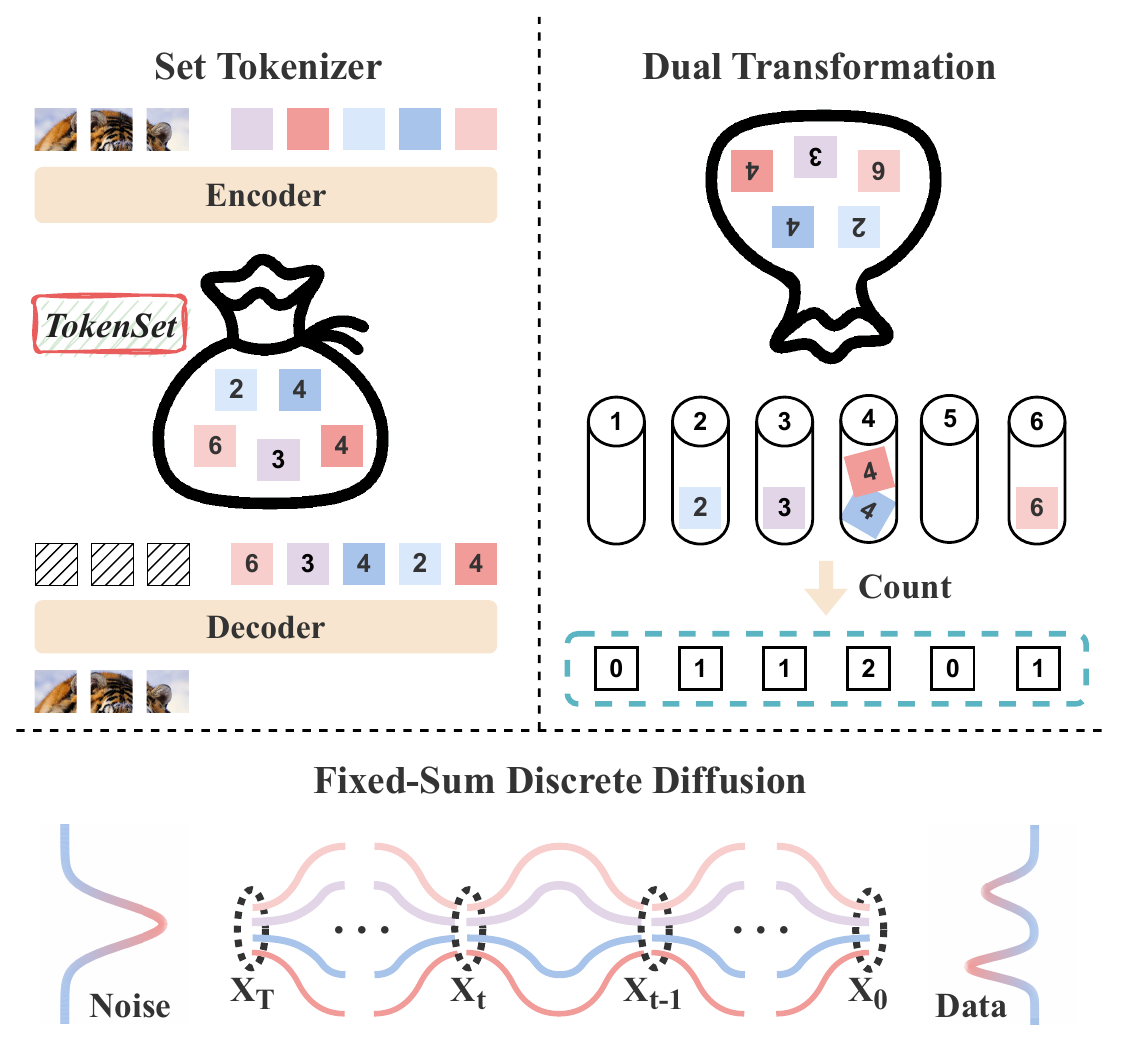}
    \vspace{-1.6em}
    \caption{Pipeline of our method.}
    \label{fig:teaser}
    \vspace{-1.0em}
\end{figure}

This paper introduces a completely novel way of visual compression and distribution modeling, \emph{TokenSet} (\cref{fig:teaser}). During the compression stage, we propose to tokenize images into unordered sets rather than position-dependent sequences. Unlike serialized tokens that maintain fixed spatial correspondence, our token set formulation enables dynamic attention allocation based on regional semantic complexity. This approach enhances global information aggregation, facilitates semantic-aware representation, and exhibits superior robustness to local perturbations.

Nevertheless, modeling set-structured data presents significant challenges compared to sequential counterparts. Existing set modeling approaches fall into two categories: The first class~\cite{zaheer2017deep,lee2019set,edwards2016towards} adopts pooling-based operations (e.g., mean/sum/max/similar operations) to extract low-dimensional features, it lacks direct supervision on each element in the set, often yielding suboptimal results; Other correspondence-based methods (e.g., DETR's Hungarian matching~\cite{carion2020end}) seek to construct element-wise supervision through bipartite matching. However, the inherent instability of dynamic matching mechanisms causes supervisory signals to vary across training iterations, leading to suboptimal convergence.

To address it, we devise a \emph{dual transformation} mechanism that converts set modeling into a sequence modeling problem. Specifically, we count the occurrences of each unique token index in the set, transforming unordered data into a structured sequence where: (i) the sequence length equals the codebook size, (ii) each element represents non-negative integer counts, (iii) the summation of all elements equals the number of elements in the set.

While existing discrete modeling approaches (e.g., VQ-Diffusion~\cite{gu2022vqdiff}) handle fixed-length integer sequences but ignore summation constraints, and continuous diffusion models~\cite{lin2024common} can preserve element sums while struggling with discrete value representations, no current approach simultaneously satisfies all three constraints. We therefore propose Fixed-Sum Discrete Diffusion models. By introducing a constant-sum prior, we simultaneously satisfy all three critical properties and achieve effective modeling of this structured data.

Our contributions can thus be summarized as:

\begin{enumerate}
    \item We propose a novel image generation paradigm through a set-based representation, fundamentally departing from conventional serialized representations.
    \item The TokenSet exhibits global contextual awareness, enabling dynamic token allocation to semantic complexity while maintaining robustness against local perturbations.
    \item We propose an effective solution for modeling discrete set data through dual transformation, establishing a bijection between unordered sets and serialized data.
    \item We propose Fixed-Sum Discrete Diffusion, a dedicated generative framework that explicitly enforces summation constraints in discrete data modeling, achieving superior modeling of set distributions.
\end{enumerate}
\section{Related Work}
\label{sec:related}

\subsection{Image tokenization} 
Image tokenization compresses images from high-dimensional pixel space into a compact representation, facilitating subsequent understanding and generation tasks. Early approaches like Variational Autoencoders (VAEs)~\cite{kingma2013auto} map input images into low-dimensional continuous latent distributions. Building on this, VQVAE~\cite{vqvae17aaron} and VQGAN~\cite{vqgan21patrick} project images into discrete token sequences, associating each image patch with an explicit discrete token. 
Subsequent works VQVAE-2~\cite{razavi19vqvae2}, RQVAE~\cite{rqvae22lee}, and MoVQ~\cite{zheng2022movq} leverage residual quantization strategies to encode images into hierarchical latent representations. Meanwhile, FSQ~\cite{mentzer23fsq}, SimVQ~\cite{zhu24simvq}, and VQGAN-LC~\cite{zhu24vqganlc} address the representation collapse problem when scaling up codebook sizes. Other innovations include a dynamic quantization strategy in DQVAE~\cite{huang23dq}, integration of semantic information in ImageFolder~\cite{li2024imagefolder}, and architectural refinements~\cite{yu21vitvqgan,cao2023efficient}. Recently, TiTok~\cite{titok24yu} explores 1D latent sequences for image representation, achieving good reconstruction at an impressive compression ratio.

Despite these advances, previous approaches predominantly encode images into token sequences, where each element corresponds strictly to fixed image positions. This paper proposes representing images as unordered token sets, thereby eliminating positional bias while effectively capturing global visual semantics.

\subsection{Set Modeling}

Early set-based representations include Bag-of-Words (BoW) \cite{salton1975bow, joachims1998bow, pang2002thumbs} and its visual counterparts \cite{sivic2003videogoogle, csurka2004visualbok, lazebnik2006beyondbow}. More recently, CoC \cite{ma2023setofpoints} proposes to treat an image as a set of points via clustering. However, these set-based representations lose information from the original data. Conversely, certain data modalities—such as point clouds and bounding boxes—inherently align with set representations. This has motivated substantial research efforts to model permutation-invariant data, yet it presents three fundamental challenges.

Firstly, prevailing generative paradigms such as auto-regressive (AR) models \cite{sun2024llamagen, tian2025var, vqgan21patrick} and diffusion models \cite{ho2020ddpm, diffbeatgan21, gu2022vqdiff, gat2025discrete, dit23peebles} are designed for sequential data modeling, making them incompatible with unordered set-structured data.

Secondly, processing permutation-invariant data necessitates strictly symmetric operations (e.g., sum, max, or similar) to avoid positional dependencies~\cite{zaheer2017deep,lee2019set,edwards2016towards}. However, this constraint prevents the use of powerful tools such as convolution and attention, thereby bottlenecking the model's capacity. 

Thirdly, effective modeling of complex data distributions typically requires per-element supervision signals, yet unordered sets inherently lack such mechanisms. Existing approaches like DSPN~\cite{zhang2019dspn} employ Chamfer loss for supervision, while TSPN~\cite{kosiorek2020tspn} and DETR~\cite{carion2020end} utilize Hungarian matching. However, these matching processes are inherently unstable and often lead to inconsistent training signals. Alternative approaches like PointCloudGAN~\cite{li2018point} attempt to directly model the global distribution, which compromises training effectiveness and overall performance.

This paper bypasses these limitations through a dual transformation operation, effectively transforming sets into sequences. This transformation enables us to leverage various sequence-based modeling methods to tackle the challenging task of set modeling.
\section{Method}
\label{sec:method}

\subsection{Image Set Tokenizer}

The key to tokenizing an image into a set is to eliminate position dependencies between visual tokens and the fixed position of the image. While prior work TiTok~\cite{titok24yu} converts images into 1D token sequences by removing 2D positional relationships, it preserves fixed 1D positional correspondence. We start from this approach and develop a completely position-agnostic tokenization framework. 
Our architecture employs Vision Transformers (ViT)~\cite{vit21alex} for both encoder and decoder components. The encoder processes image patches alongside learnable latent tokens, producing continuous latent representations that are discretized through a VQVAE~\cite{vqvae17aaron} codebook. This process generates a 1D token sequence $\mathbf{T} = [ t_1, t_2, \dots, t_M ], \text{where } t_i \in \{ 0, 1, \dots, C - 1 \}$, with $C$ denoting the codebook size and $M$ representing the token count.

\begin{figure}[tp]
    \centering
    \includegraphics[width=\linewidth]{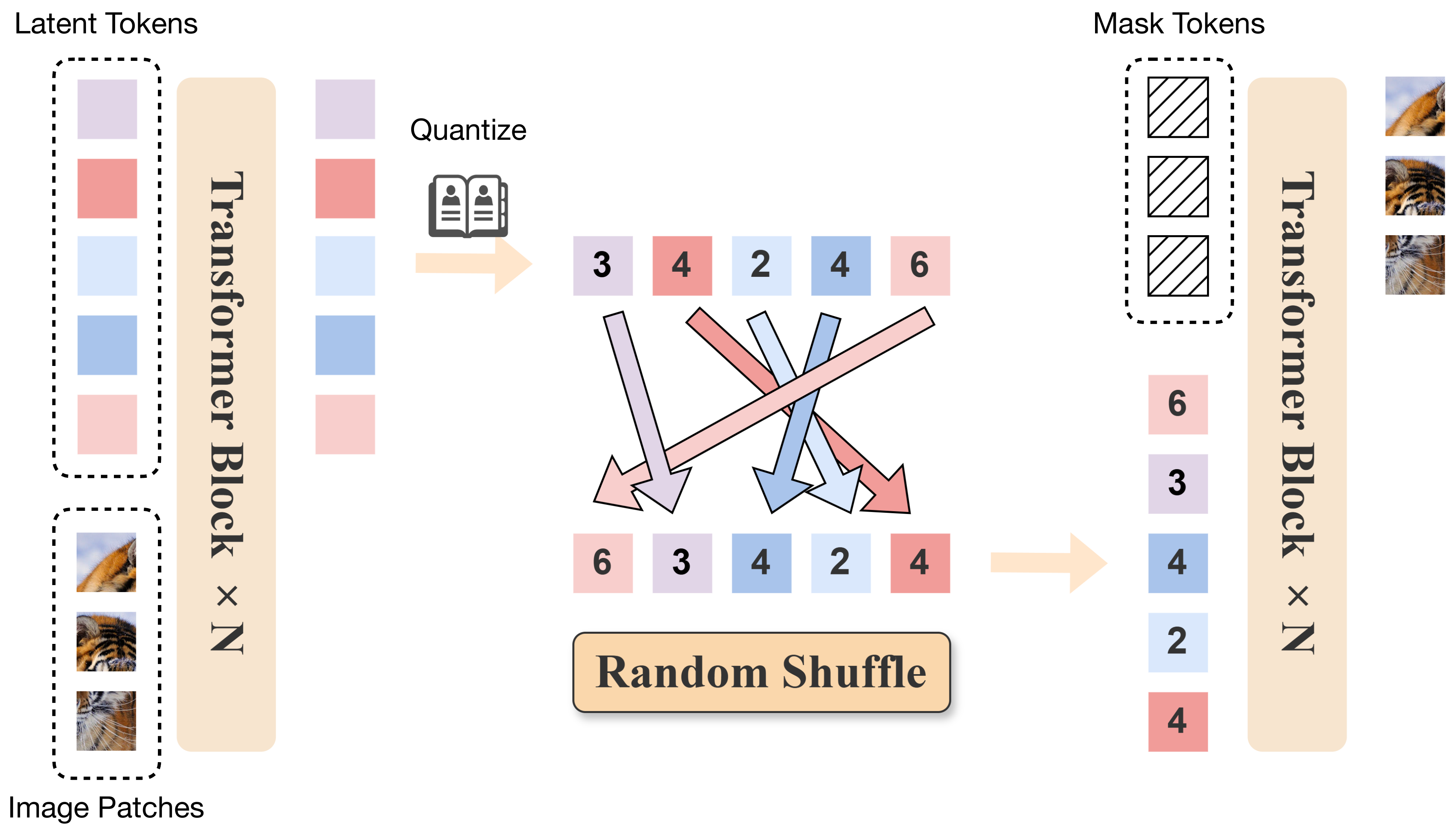}
    \vspace{-1.5em}
    \caption{Pipeline of our set tokenizer.}
    \label{fig:tokenizer}
    \vspace{-1.0em}
\end{figure}

To eliminate the 1D position bias, we introduce permutation invariance during decoding. Specifically, we define $\mathcal{T}$ as the set representation of $\mathbf{T}$ \footnote[2]{Strictly speaking, the structure should be referred to multiset due to the potential inclusion of duplicate elements.}:
\begin{equation}
  \mathcal{T} = \left\{ t_1, t_2, \dots, t_M \right\},
  \label{eq:permutation}
\end{equation}
where all permutations of $\mathbf{T}$ are considered equivalent. During training, we randomly permute tokens before decode input while maintaining reconstruction targets, as illustrated in \cref{fig:tokenizer}. Although the permutation space grows factorially with $M$, empirical results demonstrate effective learning of permutation invariance through partial permutation sampling.

This set-based tokenization demonstrates three principal advantages over serialized tokenization: First, by decoupling tokens from fixed spatial positions, the model learns to dynamically allocate tokens based on global image content rather than local patch statistics. Second, the global receptive field significantly improves noise robustness by preventing over-reliance on local features. Third, through training, tokens spontaneously develop specialized attention patterns focusing on semantically distinct regions (e.g., objects vs. textures). We empirically verify these characteristics in ~\cref{ssec:tokenizer_analysis} through quantitative evaluations and visualization studies.

\subsection{Dual transformation}

After tokenizing, an image is represented as an unordered token set $\mathcal{T}=\{ t_1, t_2, \dots, t_M \}$. Modeling such complex sets using neural networks presents significant challenges, primarily due to the inherent unordered nature of sets and the lack of effective supervision of individual elements. 

Existing sequential modeling approaches, particularly autoregressive~\cite{vqgan21patrick, sun2024llamagen} and diffusion models~\cite{diffbeatgan21, gu2022vqdiff, rombach2022sd, dit23peebles}, face inherent limitations when processing set-structured data. These methods fundamentally rely on the positional ordering of elements, making them not suitable for permutation-invariant sets where both element order ambiguity and exponential permutation possibilities exist. Alternative approaches like PointGAN~\cite{li2018point} suffer from training instability and a lack of efficient representation for permutation invariant data. Other methods like DETR~\cite{carion2020end} leverage Hungarian matching to achieve the set correspondence. However, it suffers from matching instability, hindering robust modeling. 

\begin{figure*}[tp]
    \centering
    \includegraphics[width=\linewidth]{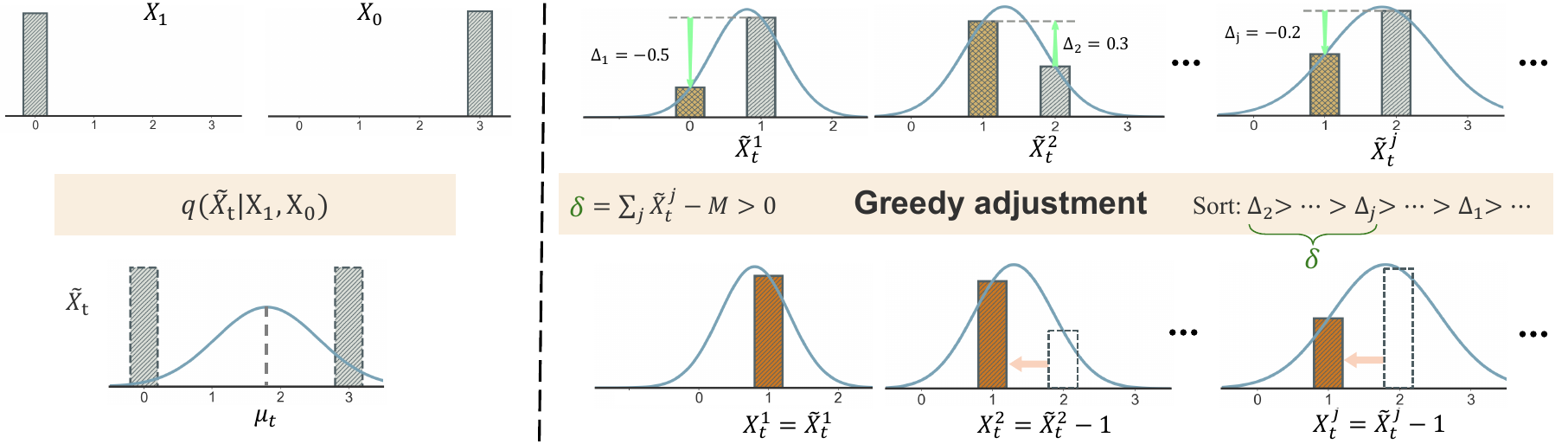}
    \vspace{-1.5em}
    \caption{Fixed-Sum Diffusion Process. Sample $X_t$ at noise level $t=0.6$ is first sampled from the mixed gaussian distribution of $X_0$ and $X_1$, and then adjusted through greedy adjustment. Samples dropped during greedy adjustment are marked with dashed line.}
    \vspace{-0.5em}
    \label{fig:diff_forward}
\end{figure*}

To address these challenges, we propose a dual transformation mechanism (\cref{fig:teaser}) that bidirectionally converts between unordered sets and structured sequences. Given a token set $\mathcal{T} = \left\{ t_1, t_2, \dots, t_M \right\}$, we construct a count vector $\mathbf{X} = (x^0, x^1, ... x^{C-1}) \in \mathbb{N}^C$ through:
\begin{equation}
  x^j = \sum_{i=1}^{M} \delta(t_i, j), \quad \text{for } j = 0, 1, \dots, C - 1,
 \label{eq:dual}
\end{equation}

where $\delta(t_i, j)$ denotes the Kronecker delta function. By doing this, we effectively convert the unordered token set into serialized data without loss of any information. Furthermore, the converted sequence data $\mathbf{X}$ has three critical structural priors:

\begin{itemize}
    \item \textbf{Fixed-length sequence}: The count vector \( \mathbf{X} \) contains $C$ elements, corresponding to the size of the codebook, ensuring a fixed length sequence.
    
    \item \textbf{Discrete count values}: Each element \( x_j \), which records codebook item frequencies, is an integer in $[0, M]$ , where $M$ is the number of tokens extracted from the encoder.
    
    \item \textbf{Fixed-sum constraint}: The summation of all values equals the number of encoded tokens:
    \begin{equation}
       \sum_{j=0}^{C - 1} x^j = M.
      \label{eq:sumconst}
    \end{equation}
\end{itemize}

In summary, the dual transformation establishes a bidirectional mapping between set and sequence representations, offering two fundamental advantages: (1) It reduces the challenging problem of modeling permutation-invariant sets to the well-studied domain of sequence modeling, crucially enabling auto-regressive and diffusion frameworks for modeling. (2) The identified structural priors—fixed sequence length, discrete value constraints, and summation conservation—provide mathematically grounded regularization that guides effective model learning. 

\subsection{Fixed-Sum Discrete Diffusion}

Given that our dual-transformed sequential data exhibits three favorable prior properties, we systematically investigated several modeling approaches. While both auto-regressive models~\cite{sun2024llamagen} and standard discrete diffusion methods~\cite{gu2022vqdiff, gat2025discrete} are effective for discrete-valued data, with the latter being particularly suitable for fixed-length sequences, they do not inherently guarantee the fixed-summation property.

Conversely, continuous diffusion models~\cite{dit23peebles} can naturally preserve both fixed-length and fixed-summation constraints through their mean-preserving MSE loss~\cite{lin2024common}, but they struggle with discrete distribution modeling~\cite{chen2022analog}.

To synergistically combine the strengths of these approaches while satisfying all three priors, 
we propose a novel modeling approach called \textbf{Fixed-Sum Discrete Diffusion} (FSDD), illustrated in~\cref{fig:diff_forward}. Inspired by continuous diffusion methods that enforce summation constraints during intermediate denoising steps, this method integrates a constrained diffusion path within a discrete flow matching architecture. The key innovation lies in ensuring that samples at every intermediate step strictly adhere to the fixed-sum constraint.

\subsubsection{Training pipeline}

\textbf{Diffusion Process} The initial noise sample $\mathbf{X}_1$ is drawn from a multinomial distribution over integer vectors of length $C$ with a fixed summation $M$:
\begin{equation}\label{eq:init_state}
\begin{aligned}
    \mathbf{X}_1 &= (x_1^0, x_1^1, \dots, x_1^{C-1}), \\
    \text{where}\quad &0 \leq x_1^j \leq M, \quad \sum_{j=0}^{C-1} x_1^j = M.
\end{aligned}
\end{equation}

Given $\mathbf{X}_1$ sampled from the noise distribution and $\mathbf{X}_0$ from the data distribution, both satisfying $\sum \mathbf{X} = M$, we define the constrained diffusion process as:
\begin{equation}
    \label{eq:diffusion}
    q(\tilde{\mathbf{X}}_t|\mathbf{X}_1, \mathbf{X}_0) = \mathcal{N}(\boldsymbol{\mu}_t,\,\boldsymbol{\sigma}_t^2).
\end{equation}

The parameters $\boldsymbol{\mu}_t$ and $\boldsymbol{\sigma}_t$ satisfy:
\begin{equation}\label{eq:mu_sigma}
    \boldsymbol{\mu}_t = t\,\mathbf{X}_1 + (1 - t)\,\mathbf{X}_0,\quad
    \boldsymbol{\sigma}_t = \frac{\lvert \mathbf{X}_1 - \mathbf{X}_0 \rvert}{4}.
\end{equation}

This design guarantees the constraint on the summation expectation:
\begin{equation}
\mathbb{E}_{\tilde{\mathbf{X}}_t} \left[ \sum \tilde{\mathbf{X}}_t \right] = M,\quad \forall\, t \in [0,1].
\end{equation}

However, we still cannot guarantee that each individual sample $\tilde{\mathbf{X}}_t$ satisfies this constraint. Therefore, we perform dynamic adjustments to ensure it.

\noindent\textbf{Greedy Adjusting}
Our adjustment protocol operates under the core objective of preserving the likelihood of $\tilde{\mathbf{X}}_t$. Specifically, if the summation of $\tilde{\mathbf{X}}_t$ exceeds $M$, we reduce certain elements of $\tilde{\mathbf{X}}_t$ through a greedy selection criterion: For each element, we quantify the likelihood degradation caused by each adjustment and prioritize adjusting those that can increase the likelihood or minimize its reduction. Thus, the adjusted sample $\mathbf{X}_t$ adheres to both the fixed-sum constraints and the probability distribution in \cref{eq:diffusion}. We provide an illustration in \cref{fig:diff_forward} and pseudo code in \cref{alg:adjust_short}. During training, we implement this greedy adjustment strategy at every diffusion step to integrate the fixed-sum constraint.

\begin{algorithm}[t]
\caption{Greedy Adjusting Sampling}
\label{alg:adjust_short}
\begin{algorithmic}[1]
\REQUIRE Target sum $M$, sample distribution $q$
\ENSURE Sample $\mathbf{X}_t$ satisfying $\sum \mathbf{X}_t = M$
\STATE Sample $\tilde{\mathbf{X}}_t\sim q(\cdot|\mathbf{X}_1,\mathbf{X}_0)$
\STATE $\delta \leftarrow \sum\tilde{\mathbf{X}}_t - M$
\STATE $\mathbf{d}\leftarrow q(\tilde{\mathbf{X}}_t-\mathrm{sgn}(\delta)\mathbf{1}|\mathbf{X}_1,\mathbf{X}_0)-q(\tilde{\mathbf{X}}_t|\mathbf{X}_1,\mathbf{X}_0)$
\STATE $\mathbf{j}^*\leftarrow\arg\mathrm{Topk}(\mathbf{d}, |\delta|)$
\STATE Initialize $\mathbf{X}_t \leftarrow \tilde{\mathbf{X}}_t$
\STATE Adjust $\mathbf{X}_t[\mathbf{j}^*]\leftarrow\mathbf{X}_t[\mathbf{j}^*]-\mathrm{sgn}(\delta)\mathbf{1}[\mathbf{j}^*]$
\RETURN $\mathbf{X}_t$
\end{algorithmic}
\end{algorithm}

Moreover, the fixed-sum discrete diffusion employs the standard discrete diffusion loss, where the denoising network $\theta$ is trained via cross-entropy loss to predict $\mathbf{X}_0$ from noisy input $\mathbf{X}_t$, maintaining discrete state transitions while preserving the summation invariant.

\subsubsection{Inference strategy}

The inference process of Fixed-Sum Diffusion follows an iterative denoising scheme with enforced summation constraints. Starting from a noise sample $\mathbf{X}_1$ that satisfies $\sum \mathbf{X}_1 = M$, we progressively refine the sample through discrete transitions:
\begin{equation}\label{eq:denoising_step}
 p_{\theta}(\mathbf{X}_{t-\Delta t}|\mathbf{X}_t)=\sum_{\mathbf{x}_0}q(\mathbf{X}_{t-\Delta t}|\mathbf{X}_t,\mathbf{X}_0)\,p_{\theta}(\mathbf{X}_0|\mathbf{X}_t),
\end{equation}

Here, $p_{\theta}(\mathbf{X}_0|\mathbf{X}_t)$ represents the discrete data distribution predicted from the noisy data. We employ the top-$p$ sampling strategy to generate $\mathbf{X}_0$ candidates, which are then processed through the posterior term $q(\tilde{\mathbf{X}}_{t-\Delta t}|\mathbf{X}_t, \mathbf{X}_0)$. This term implements a truncated Gaussian discretization:
\begin{equation}\label{eq:diffusion_step}
\begin{aligned}
    q(\tilde{\mathbf{X}}_{t-\Delta t}|\mathbf{X}_t,\mathbf{X}_0)=\mathcal{N}(\boldsymbol{\mu}_{t-\Delta t},\boldsymbol{\sigma}_{t-\Delta t}^2),
\end{aligned}
\end{equation}
with parameters defined as:
\begin{equation}\label{eq:diffusion_parameters}
\begin{aligned}
\small
\boldsymbol{\mu}_{t-\Delta t}&=\left(1-\frac{\Delta t}{t}\right)\mathbf{X}_t+\frac{\Delta t}{t}\mathbf{X}_0,\\[6pt]
\boldsymbol{\sigma}_{t-\Delta t}&=\frac{\lvert \mathbf{X}_1-\mathbf{X}_0\rvert}{4}\cdot f(t-\Delta t),
\end{aligned}
\end{equation}
where $f(\cdot)$ controls the truncation ratio during sampling. To ensure strict adherence to the summation constraint, we apply the greedy adjustment on $\tilde{\mathbf{X}}_{t-\Delta t}$ to ensure $\sum \mathbf{X}_{t-\Delta t} = M$, effectively bridging the potential gap between the training and inference phases. 
\section{Experiments}
\label{sec:exp}
\subsection{Setting}

We conducted our experiments on the ImageNet dataset~\cite{imagenet09deng}, with images at a resolution of $256\times 256$. We report our results on the 50,000-images ImageNet validation set, utilizing the Fréchet Inception Distance (FID) metric~\cite{nashequ17heusel}. Our evaluation protocol is provided by~\cite{diffbeatgan21}.

\textbf{Implementation Details.} For tokenizer training, we followed the strategy in TiTok~\cite{titok24yu} and applied data augmentations, including random cropping and horizontal flipping. We used the AdamW optimizer~\cite{Loshchilov2019adamw} with a base learning rate of $1e\text{-}4$ and a weight decay of $1e\text{-}4$. The model is trained on ImageNet for $1000k$ steps, with a batch size of $256$, equivalent to $200$ epochs. We implemented a learning rate warm-up phase followed by a cosine decay schedule, with gradient clipping at a threshold of $1.0$. An Exponential Moving Average (EMA) with a $0.999$ decay rate was adopted, and we report results from the EMA models. To enhance quality and stabilize training, we incorporated a discriminator loss~\cite{vqgan21patrick} and trained only the decoder during the final $500k$ steps. Additionally, we utilized MaskGIT's proxy code~\cite{maskgit22chang} following~\cite{titok24yu} to facilitate training.

The generator configuration aligned with DiT~\cite{dit23peebles}. We used random horizontal flipping as data augmentation. All models were optimized with AdamW~\cite{Loshchilov2019adamw} using a constant learning rate of 1e-4 and a batch size of 256, and trained for 200 epochs. We implemented EMA with a decay rate of $0.9999$ throughout the training. 
For inference, we utilized $25$ sampling steps combined with classifier-free guidance to further enhance the image quality.

\begin{figure}[t]
    \centering
    \begin{subfigure}[t]{0.15\textwidth}
        \centering
        \includegraphics[width=\textwidth]{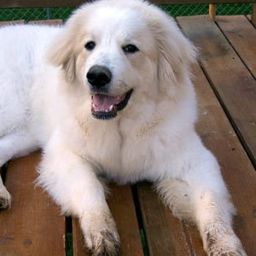}
        \caption{Original image}
    \end{subfigure}
    \hfill
    \begin{subfigure}[t]{0.15\textwidth}
        \centering
        \includegraphics[width=\textwidth]{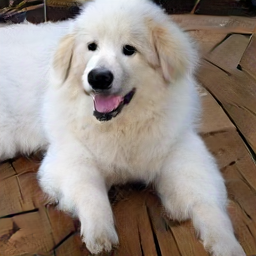}
        \caption{Original order}
    \end{subfigure}
    \hfill
    \begin{subfigure}[t]{0.15\textwidth}
        \centering
        \includegraphics[width=\textwidth]{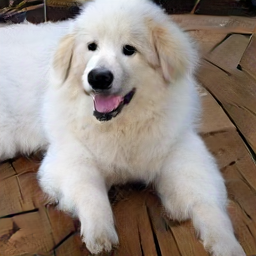}
        \caption{Reversed order}
    \end{subfigure}
    
    \vspace{0.1em} 

    \begin{subfigure}[t]{0.15\textwidth}
        \centering
        \includegraphics[width=\textwidth]{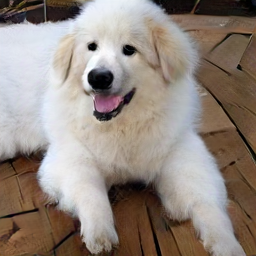}
        \caption{Random shuffle}
    \end{subfigure}
    \hfill
    \begin{subfigure}[t]{0.15\textwidth}
        \centering
        \includegraphics[width=\textwidth]{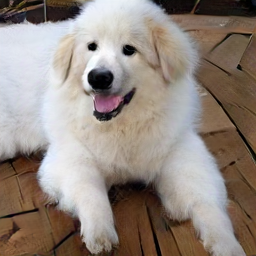}
        \caption{Sorted ascending}
    \end{subfigure}
    \hfill
    \begin{subfigure}[t]{0.15\textwidth}
        \centering
        \includegraphics[width=\textwidth]{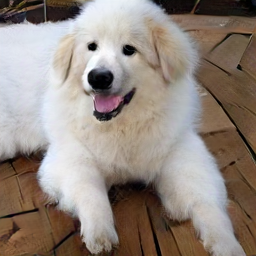}
        \caption{Sorted descending}
    \end{subfigure}
    \vspace{-.6em}
    \caption{Visual comparison of the reconstructed images from various order permutations of the encoded tokens. All reconstructed images are identical, demonstrating the set-based tokenizer is permutation-invariance.}
    \label{fig:set-image}
    \vspace{-.8em}
\end{figure}

\subsection{Set tokenizer}
\label{ssec:tokenizer_analysis}

In contrast to sequential image tokenization approaches, representing images as token sets introduces distinct properties, including permutation invariance, global context awareness, and enhanced robustness against local perturbations. Furthermore, we demonstrate that set-based tokenization can simultaneously achieve precise reconstruction while inherently organizing tokens into semantically coherent clusters.

\begin{figure*}[tp]
    \centering
    \includegraphics[width=\linewidth]{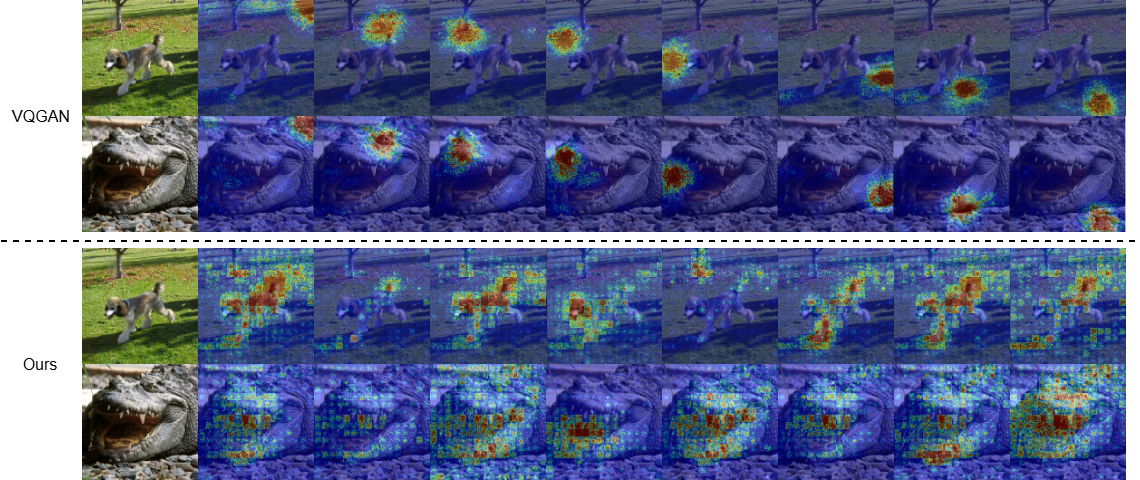}
    \vspace{-1.5em}
    \caption{The receptive fields of individual tokens. Each column represents the receptive field corresponding to the same token. Previous methods such as VQGAN~\cite{vqgan21patrick} encoded tokens strictly correspond to specific positions. In contrast, our approach demonstrates a unique property that many tokens possess global receptive fields.}
    \label{fig:global}
    \vspace{-0.7em}
\end{figure*}

\subsubsection{Permutation-invariance}
\label{permutation-set}
We test the permutation-invariance of our tokenizer by reconstructing images from encoded tokens arranged in different orders. Specifically, we decode the tokens in five different sequence orders: (1) the original order, (2) the reversed order, (3) a randomly shuffled order, (4) tokens sorted in ascending order, and (5) tokens sorted in descending order. As shown in \cref{fig:set-image}, all reconstructed images are visually identical, indicating the permutation-invariance of our tokenizer. This invariance is further supported by the quantitative results in \cref{table:set}, which are consistent across different token orders. These findings demonstrate that the network can successfully learn permutation invariance by training on only a subset of permutations.

\begin{table}[t]
    \centering
    \small
    \setlength{\tabcolsep}{3.2pt}
    \renewcommand{\arraystretch}{1.15}
    \begin{tabular}{l|ccccc}
        Order & Original & Reversed & Shuffled & Ascending & Descending \\
        \hline
        rFID$\downarrow$ & $3.62$ & $3.62$ & $3.62$  & $3.62$ & $3.62$\\
    \end{tabular}
    \vspace{-.6em}
    \caption{Quantitative results of reconstructed images using tokens in different decoding orders. The identical rFID across all orders confirms the permutation-invariance property of our tokenizer.}
    \label{table:set}
    \vspace{-1em}
\end{table}

\subsubsection{Global context awareness}
\label{subsubsec:global}

By enforcing permutation invariance, our framework decouples inter-token positional relationships, thereby eliminating sequence-induced spatial biases inherent in conventional image tokenization. This architectural design encourages each token to holistically integrate global contextual information, effectively expanding its theoretical receptive field to encompass the entire feature space. To empirically validate this phenomenon, we visualize the effective receptive field in \cref{fig:global}, which demonstrates the global attention mechanism. Notably, traditional sequence-based tokenizers, such as VQGAN~\cite{vqgan21patrick}, exhibit tight spatial coupling between tokens and fixed local regions. Conversely, our approach fundamentally eliminates positional bias and represents images through the composition of tokens with global receptive fields.

\begin{table}[t]
    \centering
    \small
    \setlength{\tabcolsep}{6pt}
    \renewcommand{\arraystretch}{1.15}
    \begin{tabular}{l | c | c c c c c}
        \multirow{2}{*}{Method} & \multirow{2}{*}{$\#$Tokens} & \multicolumn{5}{c}{Signal-to-Noise Ratio (dB)} \\
        \cline{3-7}
        & & 40 & 30 & 20 & 10 & 1 \\
        \hline
        VQGAN~\cite{vqgan21patrick} & 256 & 69.2 & 36.6 & 10.9 & 1.6 & 0.6 \\
        TiTok~\cite{titok24yu} & 128 & 83.4 & 55.4 & 21.4 & 3.8 & 1.4 \\
        TiTok~\cite{titok24yu} & 256 & 77.2 & 44.5 & 13.2 & 1.8 & 0.7 \\
        \hline
        TokenSet \dag & 128 & 89.4 & 68.1 & 38.1 & 12.7 & 6.8 \\
        TokenSet \dag & 256 & 87.0 & 62.4 & 31.1 & 9.0 & 4.3 \\
        \hline
        TokenSet & 128 & \textbf{89.5} & \textbf{68.4} & \textbf{38.9} & \textbf{13.9} & \textbf{7.9} \\
        TokenSet & 256 & 87.6 & 64.0 & 33.2 & 10.6 & 5.7 \\
    \end{tabular}
    \vspace{-.2em}
    \caption{Robustness analysis of different tokenizers under varying levels of Gaussian noise. We report the percentage (\%) of overlapping tokens between those generated from the original images and the noise-altered images. Although our method is set-based, we additionally provide results obtained by treating token sets as sequences (marked by \dag) to ensure a fair comparison.}
    \label{table:robust}
    \vspace{-.8em}
\end{table}

\begin{figure}[tp]
    \centering
    \includegraphics[width=\linewidth, trim={5 0 0 0},clip]{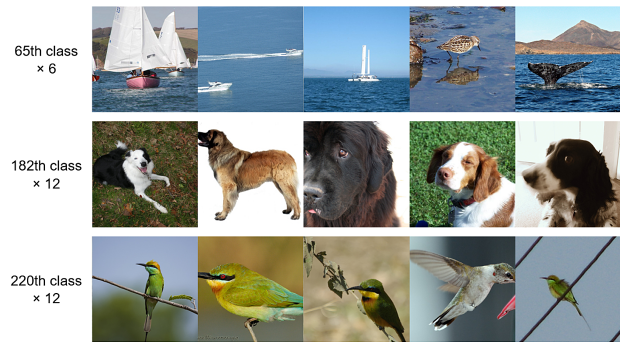}
    \vspace{-1.7em}
    \caption{Visualization of images whose encoded tokens share multiple specific classes, illustrating the inherent semantic clustering in our set-based tokenizer. Images sharing specific token classes exhibit common visual characteristics, such as marine environments, dogs, or birds.} 
    \label{fig:semantic}
    \vspace{-.1em}
\end{figure}

\begin{table}[t]
    \centering
    \small
    \setlength{\tabcolsep}{5.5pt}
    \renewcommand{\arraystretch}{1.1}
    \begin{tabular}{l | c | c c | c}
        Method & Encoder & $\#$Tokens & $\#$Codebook & Acc@1(\%) \\
        \hline
        \textcolor{gray}{MAE\dag} & \multirow{2}{*}{\textcolor{gray}{ViT-L}} & \multirow{2}{*}{-} & \multirow{2}{*}{-} & \textcolor{gray}{64.4}\\
        \textcolor{gray}{MAE} & & & & \textcolor{gray}{75.1}\\
        \hline
        \multirow{8}{*}{TokenSet} & \multirow{8}{*}{ViT-B} & 128 & 1024 & 44.8 \\
         & & 128 & 2048 & 43.1 \\
         & & 128 & 4096 & 59.7 \\
         & & 128 & 8192 & 61.0 \\
         \cline{3-5}
         & & 32 & 4096 & \textbf{66.2} \\
         & & 64 & 4096 & 64.9 \\
         & & 128 & 4096 & 59.7 \\
         & & 256 & 4096 & 47.2 \\
    \end{tabular}
    \vspace{-.4em}
    \caption{Linear probing results on ImageNet validation set. We list the reported results of a strong self-supervised method MAE~\cite{he2022masked} for reference. \dag\ denotes MAE trained for the same 200 epochs as our set tokenizer.}
    \label{tab:linear_prob}
    \vspace{-1em}
\end{table}

\subsubsection{Robustness}
Our set tokens, unbound to specific spatial positions yet capturing global image semantics, demonstrate enhanced robustness to noise. \cref{table:robust} compares the robustness of different tokenizers against Gaussian noise injected into input images. Specifically, we added Gaussian noise with varying standard deviations to images and measured the token overlap ratio between the perturbed images and the original ones. The results indicate that our tokenizer consistently achieves higher overlap ratios across all noise levels compared to other tokenizers,  such as TiTok~\cite{titok24yu} and VQGAN~\cite{vqgan21patrick}. Furthermore, while all methods experience performance degradation with increasing noise intensity, the degradation in our approach occurs at a slower rate, underscoring its superior robustness to noise.

\subsubsection{Semantic clustering}
Given $M$ tokens drawn from $C$ classes, the representation space of our set-based tokenizer is $\binom{M+C-1}{C-1}$, significantly smaller than that of a sequence-based tokenizer with a size of $C^M$. The compact representational space produces a more efficient depiction of the image space. \cref{fig:semantic} illustrates images whose encoded tokens contain certain classes. Intriguingly, we observe that these token distributions inherently exhibit semantically coherent clustering patterns. For example, images containing six tokens belonging to the 65th class consistently depict birds, while twelve 162nd tokens represent dogs. This phenomenon may suggest implications for advancing content-based image retrieval, where semantically coherent token sets could enable more robust feature indexing. Furthermore, we verify its semantic clustering capacity through linear probing. The results shown in~\cref{tab:linear_prob} indicate that even without specialized design, our tokenizer already achieves promising performance.

\subsubsection{Reconstruction quality}
In \cref{table:tokenizer-rec}, we compare the reconstruction performance of different tokenizers using the ImageNet validation set. Although the random shuffle in our method prevents the network from leveraging the inductive positional bias of images, and the benefits from the representation space of the set are drastically reduced, we find that this seemingly infeasible approach can still achieve good performance comparable to previous mainstream methods~\cite{vqgan21patrick, rqvae22lee}.

\begin{table}[t]
    \centering
    \small
    \setlength{\tabcolsep}{6pt}
    \renewcommand{\arraystretch}{1.15}
    \begin{tabular}{l|cc|c}
        Method &$\#$Tokens &$\#$Codebook & rFID$\downarrow$\\
        \hline
        Taming-VQGAN~\cite{vqgan21patrick} &256 &1024 &7.97\\
        Taming-VQGAN~\cite{vqgan21patrick} &256 &16384 &4.98\\
        RQVAE~\cite{rqvae22lee} &256 &16384 &3.20\\
        MaskGit-VQGAN~\cite{maskgit22chang} &256 &1024 &2.28\\
        ViT-VQGAN~\cite{vitvqgan22yu} &1028 &8192 &1.28\\
        TiTok~\cite{titok24yu} &64 &4096 &1.70\\
        \hline
        TokenSet &128 &2048 &3.62\\
        TokenSet &128 &4096 &2.74\\
    \end{tabular}
    \vspace{-.6em}
    \caption{Comparison of the reconstruction results of different tokenizers on the ImageNet benchmark at a resolution of $256\times256$. }
    \label{table:tokenizer-rec}
    \vspace{.3em}
\end{table}

\subsection{Fixed-sum Discrete Diffusion}

\begin{table}[t]
    \centering
    \small
    \setlength{\tabcolsep}{5.5pt}
    \renewcommand{\arraystretch}{1.15}
    \begin{tabular}{l |ccc| c}
        Method & Set &Discrete &Fixed-sum &gFID$\downarrow$\\
        \hline
        AR-order1 &&\checkmark&&6.55\\ 
        AR-order2 &&\checkmark&&6.62\\
        AR-random &\checkmark &\checkmark &&8.99\\
        SetAR &\checkmark &\checkmark &&6.92\\ 
        Discrete Diffusion &\checkmark &\checkmark &&6.23\\
        Continuous Diffusion &\checkmark &&\checkmark &75.45\\
        \hline
        FSDD &\checkmark &\checkmark &\checkmark &\textbf{5.56}\\
    \end{tabular}
    \vspace{-.7em}
    \caption{Ablation study on different modeling methods. Our tokenizer exhibits three important priors: set, discreteness, and fixed-sum constraint. We present different modeling methods, indicating which priors they satisfy and their corresponding performance. }
    \label{tab:ablation-study-final}
    \vspace{-.7em}
\end{table}

\subsubsection{Modeling through priors}

In \cref{tab:ablation-study-final}, we compare different modeling methods that leverage different priors. All experiments are conducted using models of the ViT-Small~\cite{vit21alex} scale. 

First, as described in \cref{permutation-set}, any permutation of the image token set can equivalently reconstruct the image. We therefore consider randomly sampling two different permutations for autoregressive modeling~\cite{vqgan21patrick}, termed AR-order1 and AR-order2. We find that both achieve nearly identical performance. This observation suggests that the permutation-invariant property remains underutilized. To address this, we propose training a single autoregressive model to simultaneously learn all possible permutations of the set, termed AR-random. However, this method exhibits poor generation performance due to the extreme difficulty of modeling the intractably large permutation space.

To overcome this limitation, we apply the dual transformation and subsequently model the resulting sequence distribution. We experiment with both autoregressive~\cite{vqgan21patrick} and discrete diffusion~\cite{gat2025discrete}, leveraging set and discrete properties, which perform better than the AR-random. However, they inherently fail to guarantee adherence to our fixed-sum prior, leading to mediocre results. Likewise, modeling discrete distributions by applying continuous diffusion~\cite{dit23peebles} and subsequent quantization is also proved ineffective.

In contrast, our proposed Fixed-Sum Discrete Diffusion (FSDD) approach is uniquely capable of simultaneously satisfying all these desired properties, leading to the highest performance. This finding demonstrates the importance of incorporating all the requisite priors into the modeling method to facilitate easier modeling. It validates our design choice and highlights the synergy between the tokenizer's requirements and our modeling approach.

\begin{table}[t]
    \centering
    \small
    \setlength{\tabcolsep}{12pt}
    \renewcommand{\arraystretch}{1.1}
    \begin{tabular}{cc | cc}
        $\#$Tokens &$\#$Codebook & rFID$\downarrow$ & gFID$\downarrow$\\
        \hline
        128 &1024 &6.51 &9.93\\
        128 &2048 &3.62 &7.12\\
        128 &4096 &2.74 &\textbf{5.56}\\
        128 &8192 &\textbf{2.35} &8.76\\
        \hline
        32 &4096 &5.54 &6.91\\
        64 &4096 &3.54 &6.03\\
        128 &4096 &2.74 &\textbf{5.56}\\
        256 &4096 &\textbf{2.60} &7.07\\
        512 &4096 &2.88 &9.37\\
    \end{tabular}
    \vspace{-.8em}
    \caption{Impact of token numbers and codebook sizes on reconstruction and generation performance.}
    \label{table:ablation-number}
    \vspace{-1.2em}
\end{table}

\subsubsection{Ablation studies}
Prior studies~\cite{vqgan21patrick, li2024imagefolder} have identified a critical dilemma within the two-stage ``compress-then-model" framework for image generation: Increasing the latent space capacity steadily improves reconstruction quality, but generation quality first improves and then declines. Here we rigorously investigate whether our approach can resolve the aforementioned dilemma. 

We systematically vary the number of tokens and codebook size to study their effects on both reconstruction and generation performance. For generation evaluation, we adopt a small-scale model (36M parameters) to fit the distribution. As illustrated in \cref{table:ablation-number}, we find that moderately increasing the latent dimensions enhances both reconstruction and generation quality, but exceeding this range degrades both metrics. This diverges from observations in serialized tokenization frameworks, where increasing the latent dimension improves reconstruction quality while compromising generation performance. We attribute this discrepancy to the lack of spatial correspondence between the set-based latent space and the image grid. Unlike serialized tokenization that enables coordinate-specific mapping through spatial priors, the set-based decoder cannot establish direct spatial correspondence mechanisms. This parallels the challenges faced by diffusion models in modeling excessively intricate latent distributions, thus both approaches suffer performance degradation with excessive dimensionality increases. Crucially, these observations suggest a potential solution to the reconstruction-generation dilemma: \emph{by removing the decoder's reliance on low-effort shortcut mappings, we can align its behavior more closely with the distribution learning process of the second-stage modeling}.

Our analysis further reveals that scaling the model size yields consistent performance gains, as empirically demonstrated in \cref{table:modeling}. While these results suggest potential benefits from further model expansion, practical constraints limited our exploration beyond the current experimental scope. We leave this for future work.

\begin{table}[t]
    \centering
    \small
    \setlength{\tabcolsep}{3pt}
    \renewcommand{\arraystretch}{1.15}
    \begin{tabular}{l|cc|c|c}
        Method & $\#$Tokens & $\#$Codebook & $\#$Params & gFID$\downarrow$ \\
        \hline
        VQGAN~\cite{vqgan21patrick} & 256 & 1024 & 1.4B & 15.78 \\
        VQ-Diffusion~\cite{gu2022vqdiff} & 1024 & 2886 & 370M & 11.89 \\
        MaskGiT~\cite{maskgit22chang} & 256 & 1024 & 227M & 6.18 \\
        LlamaGen~\cite{sun2024llamagen} & 256 & 16384 & 111M & 5.46 \\
        LlamaGen~\cite{sun2024llamagen} & 256 & 16384 & 3.1B & 2.18 \\
        TiTok~\cite{titok24yu} & 128 & 4096 & 287M & 1.97 \\
        \hline
        TokenSet-S & 128 & 4096 & 36M & 5.56 \\
        TokenSet-B & 128 & 4096 & 137M & 5.09 \\
    \end{tabular}
    \vspace{-.7em}
    \caption{Modeling performance comparison on the ImageNet benchmark at $256\times256$ resolution.}
    \vspace{-1.3em}
    \label{table:modeling}
\end{table}
\section{Conclusion}
\label{sec:conclusion}

This work challenges the conventional paradigm of serialized visual representation by introducing \emph{TokenSet}, a set-based framework that dynamically allocates representational capacity to semantically diverse image regions. Through dual transformation, we establish a bijective mapping between unordered token sets and structured integer sequences, enabling effective modeling of set distributions via our proposed fixed-sum discrete diffusion. Experiments demonstrate that this approach not only achieves dynamic token allocation aligned with regional complexity, but also enhances robustness against local perturbations. By enforcing summation constraints during both training and inference, our framework resolves critical limitations in existing discrete diffusion models while outperforming fixed-length sequence baselines. 

Tokenizing image as a set offers distinct advantages over conventional sequential tokenization, introducing novel possibilities for both image representation and generation. This paradigm shift inspires new perspectives on developing next-generation generative models. In future work, we plan to conduct a rigorous analysis to unlock the full potential of this representation and modeling approach.

{
    \small
    \bibliographystyle{ieeenat_fullname}
    \bibliography{main}

\begin{thebibliography}{46}
\providecommand{\natexlab}[1]{#1}
\providecommand{\url}[1]{\texttt{#1}}
\expandafter\ifx\csname urlstyle\endcsname\relax
  \providecommand{\doi}[1]{doi: #1}\else
  \providecommand{\doi}{doi: \begingroup \urlstyle{rm}\Url}\fi

\bibitem[Cao et~al.(2023)Cao, Yin, Huang, Liu, Zhao, Zhao, and Huang]{cao2023efficient}
Shiyue Cao, Yueqin Yin, Lianghua Huang, Yu Liu, Xin Zhao, Deli Zhao, and Kaigi Huang.
\newblock Efficient-vqgan: Towards high-resolution image generation with efficient vision transformers.
\newblock In \emph{ICCV}, pages 7368--7377, 2023.

\bibitem[Carion et~al.(2020)Carion, Massa, Synnaeve, Usunier, Kirillov, and Zagoruyko]{carion2020end}
Nicolas Carion, Francisco Massa, Gabriel Synnaeve, Nicolas Usunier, Alexander Kirillov, and Sergey Zagoruyko.
\newblock End-to-end object detection with transformers.
\newblock In \emph{ECCV}, 2020.

\bibitem[Chang et~al.(2022)Chang, Zhang, Jiang, Liu, and Freeman]{maskgit22chang}
Huiwen Chang, Han Zhang, Lu Jiang, Ce Liu, and William~T. Freeman.
\newblock Maskgit: Masked generative image transformer.
\newblock In \emph{CVPR}, pages 11305--11315, 2022.

\bibitem[Chen et~al.(2022)Chen, Zhang, and Hinton]{chen2022analog}
Ting Chen, Ruixiang Zhang, and Geoffrey Hinton.
\newblock Analog bits: Generating discrete data using diffusion models with self-conditioning.
\newblock \emph{arXiv preprint arXiv:2208.04202}, 2022.

\bibitem[Csurka et~al.(2004)Csurka, Dance, Fan, Willamowski, and Bray]{csurka2004visualbok}
Gabriella Csurka, Christopher Dance, Lixin Fan, Jutta Willamowski, and C{\'e}dric Bray.
\newblock Visual categorization with bags of keypoints.
\newblock In \emph{Workshop on statistical learning in computer vision, ECCV}, pages 1--2. Prague, 2004.

\bibitem[Deng et~al.(2009)Deng, Dong, Socher, Li, Li, and Fei{-}Fei]{imagenet09deng}
Jia Deng, Wei Dong, Richard Socher, Li{-}Jia Li, Kai Li, and Li Fei{-}Fei.
\newblock Imagenet: {A} large-scale hierarchical image database.
\newblock In \emph{CVPR}, pages 248--255, 2009.

\bibitem[Dhariwal and Nichol(2021)]{diffbeatgan21}
Prafulla Dhariwal and Alexander~Quinn Nichol.
\newblock Diffusion models beat gans on image synthesis.
\newblock In \emph{Neurips}, pages 8780--8794, 2021.

\bibitem[Dosovitskiy et~al.(2021)Dosovitskiy, Beyer, Kolesnikov, Weissenborn, Zhai, Unterthiner, Dehghani, Minderer, Heigold, Gelly, Uszkoreit, and Houlsby]{vit21alex}
Alexey Dosovitskiy, Lucas Beyer, Alexander Kolesnikov, Dirk Weissenborn, Xiaohua Zhai, Thomas Unterthiner, Mostafa Dehghani, Matthias Minderer, Georg Heigold, Sylvain Gelly, Jakob Uszkoreit, and Neil Houlsby.
\newblock An image is worth 16x16 words: Transformers for image recognition at scale.
\newblock In \emph{ICLR}, 2021.

\bibitem[Edwards and Storkey(2016)]{edwards2016towards}
Harrison Edwards and Amos Storkey.
\newblock Towards a neural statistician.
\newblock \emph{ICLR}, 2016.

\bibitem[Esser et~al.(2021)Esser, Rombach, and Ommer]{vqgan21patrick}
Patrick Esser, Robin Rombach, and Bj{\"{o}}rn Ommer.
\newblock Taming transformers for high-resolution image synthesis.
\newblock In \emph{CVPR}, pages 12873--12883, 2021.

\bibitem[Gat et~al.(2025)Gat, Remez, Shaul, Kreuk, Chen, Synnaeve, Adi, and Lipman]{gat2025discrete}
Itai Gat, Tal Remez, Neta Shaul, Felix Kreuk, Ricky~TQ Chen, Gabriel Synnaeve, Yossi Adi, and Yaron Lipman.
\newblock Discrete flow matching.
\newblock \emph{Neurips}, 37:\penalty0 133345--133385, 2025.

\bibitem[Gu et~al.(2022)Gu, Chen, Bao, Wen, Zhang, Chen, Yuan, and Guo]{gu2022vqdiff}
Shuyang Gu, Dong Chen, Jianmin Bao, Fang Wen, Bo Zhang, Dongdong Chen, Lu Yuan, and Baining Guo.
\newblock Vector quantized diffusion model for text-to-image synthesis.
\newblock In \emph{CVPR}, pages 10696--10706, 2022.

\bibitem[He et~al.(2022)He, Chen, Xie, Li, Doll{\'a}r, and Girshick]{he2022masked}
Kaiming He, Xinlei Chen, Saining Xie, Yanghao Li, Piotr Doll{\'a}r, and Ross Girshick.
\newblock Masked autoencoders are scalable vision learners.
\newblock In \emph{CVPR}, 2022.

\bibitem[Heusel et~al.(2017)Heusel, Ramsauer, Unterthiner, Nessler, and Hochreiter]{nashequ17heusel}
Martin Heusel, Hubert Ramsauer, Thomas Unterthiner, Bernhard Nessler, and Sepp Hochreiter.
\newblock Gans trained by a two time-scale update rule converge to a local nash equilibrium.
\newblock In \emph{Neurips}, pages 6626--6637, 2017.

\bibitem[Ho et~al.(2020)Ho, Jain, and Abbeel]{ho2020ddpm}
Jonathan Ho, Ajay Jain, and Pieter Abbeel.
\newblock Denoising diffusion probabilistic models.
\newblock \emph{Neurips}, 33:\penalty0 6840--6851, 2020.

\bibitem[Huang et~al.(2023)Huang, Mao, Chen, and Zhang]{huang23dq}
Mengqi Huang, Zhendong Mao, Zhuowei Chen, and Yongdong Zhang.
\newblock Towards accurate image coding: {Improved} autoregressive image generation with dynamic vector quantization.
\newblock In \emph{CVPR}, pages 22596--22605, 2023.

\bibitem[Joachims(1998)]{joachims1998bow}
Thorsten Joachims.
\newblock Text categorization with support vector machines: Learning with many relevant features.
\newblock In \emph{ECML}, pages 137--142. Springer, 1998.

\bibitem[Kingma et~al.(2013)Kingma, Welling, et~al.]{kingma2013auto}
Diederik~P Kingma, Max Welling, et~al.
\newblock Auto-encoding variational bayes, 2013.

\bibitem[Kosiorek et~al.(2020)Kosiorek, Kim, and Rezende]{kosiorek2020tspn}
Adam~R Kosiorek, Hyunjik Kim, and Danilo~J Rezende.
\newblock Conditional set generation with transformers.
\newblock \emph{arXiv preprint arXiv:2006.16841}, 2020.

\bibitem[Lazebnik et~al.(2006)Lazebnik, Schmid, and Ponce]{lazebnik2006beyondbow}
Svetlana Lazebnik, Cordelia Schmid, and Jean Ponce.
\newblock Beyond bags of features: Spatial pyramid matching for recognizing natural scene categories.
\newblock In \emph{CVPR}, pages 2169--2178. IEEE, 2006.

\bibitem[Lee et~al.(2022)Lee, Kim, Kim, Cho, and Han]{rqvae22lee}
Doyup Lee, Chiheon Kim, Saehoon Kim, Minsu Cho, and Wook{-}Shin Han.
\newblock Autoregressive image generation using residual quantization.
\newblock In \emph{CVPR}, pages 11513--11522, 2022.

\bibitem[Lee et~al.(2019)Lee, Lee, Kim, Kosiorek, Choi, and Teh]{lee2019set}
Juho Lee, Yoonho Lee, Jungtaek Kim, Adam~R Kosiorek, Seungjin Choi, and Yee~Whye Teh.
\newblock Set transformer.
\newblock In \emph{ICML}, 2019.

\bibitem[Li et~al.(2018)Li, Zaheer, Zhang, Poczos, and Salakhutdinov]{li2018point}
Chun-Liang Li, Manzil Zaheer, Yang Zhang, Barnabas Poczos, and Ruslan Salakhutdinov.
\newblock Point cloud gan.
\newblock \emph{arXiv preprint arXiv:1810.05795}, 2018.

\bibitem[Li et~al.(2024{\natexlab{a}})Li, Tian, Li, Deng, and He]{mar24li}
Tianhong Li, Yonglong Tian, He Li, Mingyang Deng, and Kaiming He.
\newblock Autoregressive image generation without vector quantization.
\newblock In \emph{Neurips}, 2024{\natexlab{a}}.

\bibitem[Li et~al.(2024{\natexlab{b}})Li, Qiu, Chen, Kuen, Gu, Raj, and Lin]{li2024imagefolder}
Xiang Li, Kai Qiu, Hao Chen, Jason Kuen, Jiuxiang Gu, Bhiksha Raj, and Zhe Lin.
\newblock Imagefolder: Autoregressive image generation with folded tokens.
\newblock \emph{arXiv preprint arXiv:2410.01756}, 2024{\natexlab{b}}.

\bibitem[Lin et~al.(2024)Lin, Liu, Li, and Yang]{lin2024common}
Shanchuan Lin, Bingchen Liu, Jiashi Li, and Xiao Yang.
\newblock Common diffusion noise schedules and sample steps are flawed.
\newblock In \emph{WACV}, pages 5404--5411, 2024.

\bibitem[Loshchilov and Hutter(2019)]{Loshchilov2019adamw}
Ilya Loshchilov and Frank Hutter.
\newblock Decoupled weight decay regularization.
\newblock In \emph{ICLR}, 2019.

\bibitem[Ma et~al.(2023)Ma, Zhou, Wang, Qin, Sun, Liu, and Fu]{ma2023setofpoints}
Xu Ma, Yuqian Zhou, Huan Wang, Can Qin, Bin Sun, Chang Liu, and Yun Fu.
\newblock Image as set of points.
\newblock \emph{arXiv preprint arXiv:2303.01494}, 2023.

\bibitem[Mentzer et~al.(2023)Mentzer, Minnen, Agustsson, and Tschannen]{mentzer23fsq}
Fabian Mentzer, David Minnen, Eirikur Agustsson, and Michael Tschannen.
\newblock Finite scalar quantization: {Vq}-vae made simple.
\newblock \emph{arXiv preprint arXiv:2309.15505}, 2023.

\bibitem[Pang et~al.(2002)Pang, Lee, and Vaithyanathan]{pang2002thumbs}
Bo Pang, Lillian Lee, and Shivakumar Vaithyanathan.
\newblock Thumbs up? sentiment classification using machine learning techniques.
\newblock \emph{arXiv preprint cs/0205070}, 2002.

\bibitem[Peebles and Xie(2023)]{dit23peebles}
William Peebles and Saining Xie.
\newblock Scalable diffusion models with transformers.
\newblock In \emph{ICCV}, pages 4172--4182, 2023.

\bibitem[Razavi et~al.(2019)Razavi, Van~den Oord, and Vinyals]{razavi19vqvae2}
Ali Razavi, Aaron Van~den Oord, and Oriol Vinyals.
\newblock Generating diverse high-fidelity images with vq-vae-2.
\newblock \emph{Neurips}, 32, 2019.

\bibitem[Rombach et~al.(2022)Rombach, Blattmann, Lorenz, Esser, and Ommer]{rombach2022sd}
Robin Rombach, Andreas Blattmann, Dominik Lorenz, Patrick Esser, and Bj{\"o}rn Ommer.
\newblock High-resolution image synthesis with latent diffusion models.
\newblock In \emph{CVPR}, pages 10684--10695, 2022.

\bibitem[Salton et~al.(1975)Salton, Wong, and Yang]{salton1975bow}
Gerard Salton, Anita Wong, and Chung-Shu Yang.
\newblock A vector space model for automatic indexing.
\newblock \emph{Communications of the ACM}, 18\penalty0 (11):\penalty0 613--620, 1975.

\bibitem[Sivic and Zisserman(2003)]{sivic2003videogoogle}
Sivic and Zisserman.
\newblock Video google: A text retrieval approach to object matching in videos.
\newblock In \emph{ICCV}, pages 1470--1477. IEEE, 2003.

\bibitem[Sun et~al.(2024)Sun, Jiang, Chen, Zhang, Peng, Luo, and Yuan]{sun2024llamagen}
Peize Sun, Yi Jiang, Shoufa Chen, Shilong Zhang, Bingyue Peng, Ping Luo, and Zehuan Yuan.
\newblock Autoregressive model beats diffusion: Llama for scalable image generation.
\newblock \emph{arXiv preprint arXiv:2406.06525}, 2024.

\bibitem[Tian et~al.(2025)Tian, Jiang, Yuan, Peng, and Wang]{tian2025var}
Keyu Tian, Yi Jiang, Zehuan Yuan, Bingyue Peng, and Liwei Wang.
\newblock Visual autoregressive modeling: Scalable image generation via next-scale prediction.
\newblock \emph{Neurips}, 37:\penalty0 84839--84865, 2025.

\bibitem[van~den Oord et~al.(2017)van~den Oord, Vinyals, and Kavukcuoglu]{vqvae17aaron}
A{\"{a}}ron van~den Oord, Oriol Vinyals, and Koray Kavukcuoglu.
\newblock Neural discrete representation learning.
\newblock In \emph{Neurips}, pages 6306--6315, 2017.

\bibitem[Yu et~al.(2021)Yu, Li, Koh, Zhang, Pang, Qin, Ku, Xu, Baldridge, and Wu]{yu21vitvqgan}
Jiahui Yu, Xin Li, Jing~Yu Koh, Han Zhang, Ruoming Pang, James Qin, Alexander Ku, Yuanzhong Xu, Jason Baldridge, and Yonghui Wu.
\newblock Vector-quantized image modeling with improved vqgan.
\newblock \emph{arXiv preprint arXiv:2110.04627}, 2021.

\bibitem[Yu et~al.(2022)Yu, Li, Koh, Zhang, Pang, Qin, Ku, Xu, Baldridge, and Wu]{vitvqgan22yu}
Jiahui Yu, Xin Li, Jing~Yu Koh, Han Zhang, Ruoming Pang, James Qin, Alexander Ku, Yuanzhong Xu, Jason Baldridge, and Yonghui Wu.
\newblock Vector-quantized image modeling with improved {VQGAN}.
\newblock In \emph{ICLR}, 2022.

\bibitem[Yu et~al.(2024)Yu, Weber, Deng, Shen, Cremers, and Chen]{titok24yu}
Qihang Yu, Mark Weber, Xueqing Deng, Xiaohui Shen, Daniel Cremers, and Liang{-}Chieh Chen.
\newblock An image is worth 32 tokens for reconstruction and generation.
\newblock \emph{CoRR}, abs/2406.07550, 2024.

\bibitem[Zaheer et~al.(2017)Zaheer, Kottur, Ravanbakhsh, Poczos, Salakhutdinov, and Smola]{zaheer2017deep}
Manzil Zaheer, Satwik Kottur, Siamak Ravanbakhsh, Barnabas Poczos, Russ~R Salakhutdinov, and Alexander~J Smola.
\newblock Deep sets.
\newblock \emph{NeurIPS}, 30, 2017.

\bibitem[Zhang et~al.(2019)Zhang, Hare, and Prugel-Bennett]{zhang2019dspn}
Yan Zhang, Jonathon Hare, and Adam Prugel-Bennett.
\newblock Deep set prediction networks.
\newblock \emph{Neurips}, 32, 2019.

\bibitem[Zheng et~al.(2022)Zheng, Vuong, Cai, and Phung]{zheng2022movq}
Chuanxia Zheng, Tung-Long Vuong, Jianfei Cai, and Dinh Phung.
\newblock Movq: Modulating quantized vectors for high-fidelity image generation.
\newblock \emph{Neurips}, 35:\penalty0 23412--23425, 2022.

\bibitem[Zhu et~al.(2024{\natexlab{a}})Zhu, Wei, Lu, and Chen]{zhu24vqganlc}
Lei Zhu, Fangyun Wei, Yanye Lu, and Dong Chen.
\newblock Scaling the codebook size of vqgan to 100,000 with a utilization rate of 99\%.
\newblock \emph{arXiv preprint arXiv:2406.11837}, 2024{\natexlab{a}}.

\bibitem[Zhu et~al.(2024{\natexlab{b}})Zhu, Li, Xin, and Xu]{zhu24simvq}
Yongxin Zhu, Bocheng Li, Yifei Xin, and Linli Xu.
\newblock Addressing representation collapse in vector quantized models with one linear layer.
\newblock \emph{arXiv preprint arXiv:2411.02038}, 2024{\natexlab{b}}.

\end{thebibliography}
}

\end{document}